\documentclass[11pt]{article}
\usepackage[preprint]{acl}

\usepackage{times}
\usepackage{latexsym}
\usepackage[T1]{fontenc}
\usepackage[utf8]{inputenc}
\usepackage{microtype}
\usepackage{inconsolata}
\usepackage{graphicx}
\usepackage{amsmath}
\usepackage{amssymb}
\usepackage{booktabs}
\usepackage{multirow}
\usepackage{placeins}
\usepackage{xurl}
\usepackage{pgfplots}
\pgfplotsset{compat=1.18}

\newcommand{\correspondingauthor}{\textsuperscript{\ensuremath{\dagger}}}

\title{SkillCore: Internalizing Generalizable Skills via Scaffolding-Removed Fine-Tuning}
\author{
  \textbf{Yi Liu\textsuperscript{1}},
  \textbf{Changqing Ai\textsuperscript{2}},
  \textbf{Peijie Dong\textsuperscript{3}},
  \textbf{Xiankun Lin\textsuperscript{2}},
  \textbf{Yongkang Yang\textsuperscript{2}},\\
  \textbf{Zhezheng Hao\textsuperscript{4}},
  \textbf{Xilin Xia\textsuperscript{2}},
  \textbf{Hande Dong\textsuperscript{2}\correspondingauthor},
  \textbf{Qiang Lin\textsuperscript{2}},
  \textbf{Qiang Xu\textsuperscript{1}\correspondingauthor}\\
  \textsuperscript{1}The Chinese University of Hong Kong,
  \textsuperscript{2}Tencent, WorkBuddy Team,\\
  \textsuperscript{3}The Hong Kong University of Science and Technology (Guangzhou),
  \textsuperscript{4}Zhejiang University
}
\hypersetup{
  pdftitle={SkillCore: Internalizing Generalizable Skills via Scaffolding-Removed Fine-Tuning},
  pdfauthor={Yi Liu, Changqing Ai, Peijie Dong, Xiankun Lin, Yongkang Yang, Zhezheng Hao, Xilin Xia, Hande Dong, Qiang Lin, Qiang Xu}
}

\begin{document}
\maketitle
\begin{NoHyper}
\begingroup
\renewcommand{\thefootnote}{\ensuremath{\dagger}}
\footnotetext{Corresponding authors.}
\endgroup
\end{NoHyper}

% !TEX root = ../AnonymousSubmission2027.tex
\begin{abstract}
Runtime skills can improve agents' problem-solving capabilities by providing procedural guidance for tasks such as code review and debugging. However, these benefits reach only a minority of users, as most do not proactively invoke skills in practice. We argue that such generalizable procedures should become part of the model's capabilities rather than remain optional external add-ons, making them directly available without explicit skill loading.
To this end, we present \textbf{SkillCore}, a data pipeline and supervised fine-tuning recipe for internalizing generalizable skills. We transform real agent trajectories by removing skill-loading scaffolding while preserving the task context and subsequent agent actions, thereby supervising direct execution without external skill content. To preserve access to task-specific skills that provide genuinely external information, we mix these scaffold-removed trajectories with a small proportion of skill-native trajectories that preserve supervision for skill invocation. On a held-out skill-free evaluation, SkillCore improves task completion quality over the base Qwen3-8B and Qwen2.5-72B models by $54.4\%$ and $24.8\%$, respectively, while preserving task-specific skill invocation on a separate probe and general coding performance on SWE-bench Verified.
\end{abstract}

% !TEX root = ../AnonymousSubmission2027.tex
\section{Introduction}

\begin{figure}
    \centering
    \includegraphics[width=0.95\linewidth]{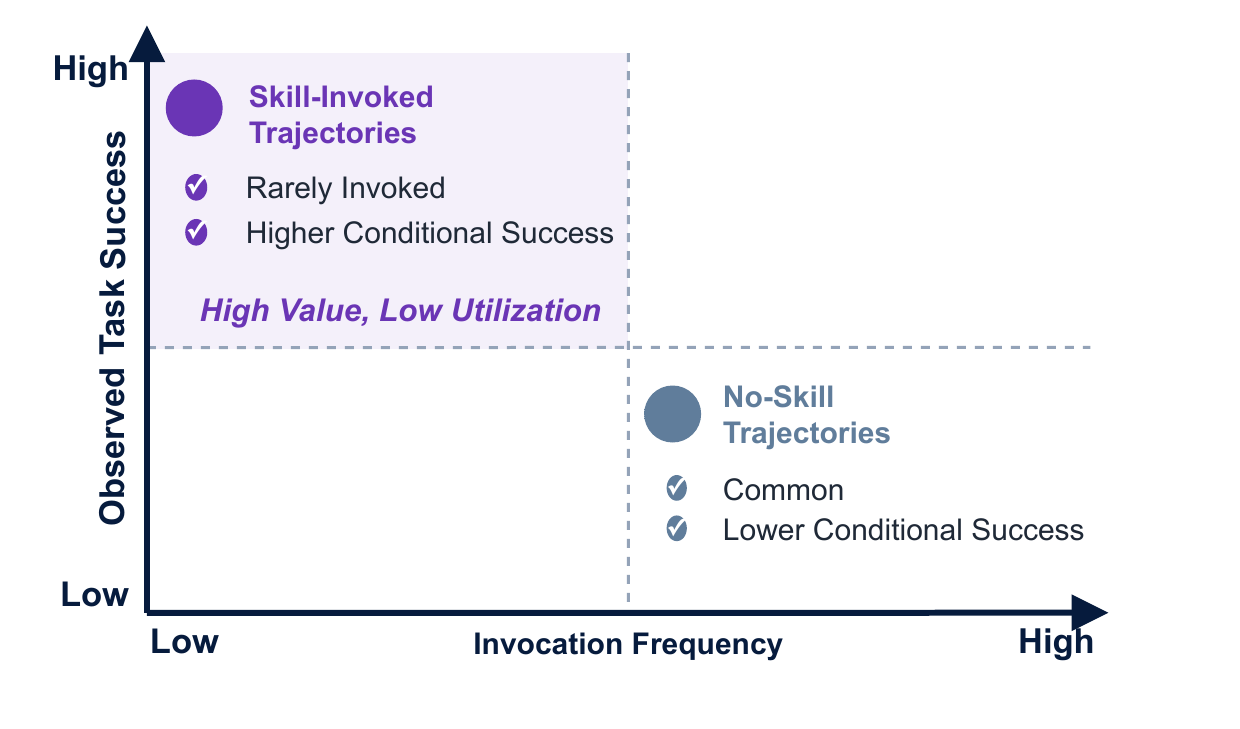}
    \caption{\textbf{The skill utilization gap.} Skill-invoked trajectories are rare but associated with higher observed task success than those without skill use. This high-value, low-utilization pattern motivates internalizing reusable procedures from generalizable skills so their benefits are available without runtime invocation.}
    \label{fig:skill-utilization-gap}
\end{figure}

% \begin{figure*}[t]
% \centering
% \includegraphics[width=0.80\textwidth]{Figures/fig1_internalization.pdf}
% \caption{\textcolor{red}{SkillCore internalizes reusable procedural capabilities by removing generalizable-skill scaffolding while preserving the demonstrated actions. Skill-native trajectories retain task-specific skill use, enabling the model to continue seeking genuinely external information when needed.}}
% \label{fig:overview}
% \end{figure*}

\begin{figure*}[t]
    \centering
    \includegraphics[width=1\linewidth]{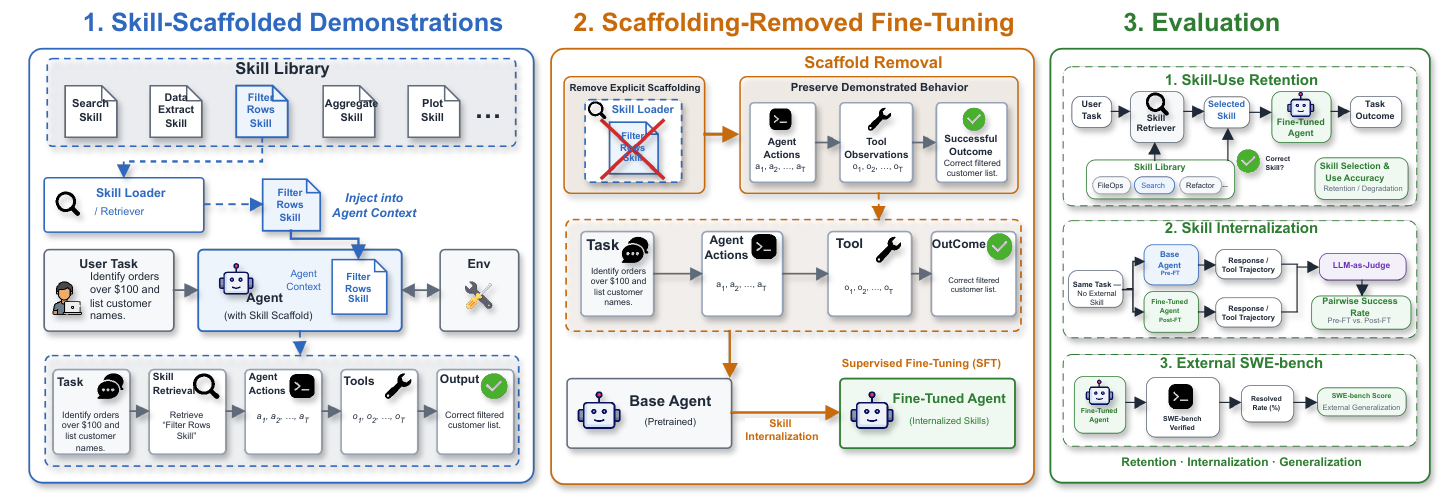}
    \caption{\textbf{Overview of SkillCore.}
\textbf{(1) Skill-Scaffolded Demonstrations:} Given a user task, a skill
retriever selects and loads a relevant skill (e.g., \emph{Filter Rows Skill})
into the agent's context. Guided by the skill, the agent interacts with the
environment to produce a successful trajectory of agent actions and
environment observations.
\textbf{(2) Scaffolding-Removed Fine-Tuning:} We remove the skill scaffold
while preserving the task context and subsequent agent actions, yielding
scaffold-removed trajectories for supervised fine-tuning that transfer the
demonstrated procedure into the model's parameters.
\textbf{(3) Evaluation:} We assess \emph{generalizable skill internalization}
(base vs.\ fine-tuned agent on the same task without external skill content,
judged by an LLM), \emph{task-specific skill invocation retention}
(appropriate skill invocation when task-specific skill content is not provided
in context but remains accessible at runtime), and \emph{preservation of general
coding capability} (the auxiliary SWE-bench Verified evaluation).}
\label{fig:overview}
\end{figure*}

Large language models (LLMs) are increasingly deployed as agents that reason, plan, and act across domains including software engineering and scientific discovery~\citep{yao2023react,wang2024agentsurvey,yang2024sweagent,boiko2023coscientist}. Their effectiveness depends not only on the underlying model but also on the surrounding runtime scaffold, including prompts, tools, memory, and control flow~\citep{sumers2024coala,schluntz2024agents}. A prominent example is the agent skill: a reusable package of instructions, scripts, and resources that is loaded into context on demand to guide a specialized workflow without updating model parameters~\citep{anthropic2025skills}. By supplying reusable procedural guidance, skills can improve agent performance across a range of tasks~\citep{wang2023voyager}.

Despite these benefits, runtime skills can benefit agents only when actually invoked. In our collected agent trajectories, only $3.2\%$ contain a successful skill load followed by subsequent agent actions, suggesting that skills are rarely used in practice. As a result, the procedural guidance they provide is not available to the agent by default. This gap motivates skill internalization: incorporating reusable skill-provided procedures into the model's parameters so that the agent can execute them without access to the corresponding skill content. The key question is therefore which skill-provided capabilities should be internalized and which should remain available through runtime invocation.

To determine what should be internalized, we distinguish between \emph{generalizable} and \emph{task-specific} skills. Generalizable skills encode reusable procedures for tasks such as systematic debugging, code review, and table extraction that transfer across tasks and environments. Task-specific skills, by contrast, contain procedures or resources tailored to a particular task or system, such as private APIs, system-bound instructions, or task-specific information. % Because their utility transfers less readily across tasks, internalizing them offers limited benefit; some also provide external information or functionality that cannot be recovered from the model's weights.
Accordingly, an effective internalization approach should make generalizable procedures available without runtime skill access while preserving runtime access to task-specific skills when specialized procedures, external information, or functionality are required.

In this work, we propose SkillCore, a framework that integrates a trajectory construction pipeline with a tailored supervised fine-tuning recipe to internalize generalizable skills while preserving runtime access to task-specific skills. Specifically, for generalizable skills, the data pipeline applies skill-scaffold removal: it removes the skill invocation, returned document, and associated loading artifacts while preserving the task context and the agent's subsequent actions. This produces training trajectories that retain the demonstrated actions while excluding the skill-relevant context that originally guided them. Fine-tuning on these trajectories transfers the demonstrated procedure into the model's parameters, enabling direct execution without the skill. By contrast, for task-specific skills, the training recipe retains a small proportion of skill-native trajectories with their invocation scaffolding intact, preserving supervision for when and how such skills should be invoked.

We evaluate SkillCore along three complementary dimensions: internalization of generalizable skills, retention of task-specific skill invocation, and preservation of general coding capability. First, a held-out skill-free evaluation assesses whether SkillCore improves model performance when generalizable skill content is unavailable. In this setting, SkillCore improves task completion quality over the base Qwen3-8B and Qwen2.5-72B models by $54.4\%$ and $24.8\%$, respectively. Second, on a separate evaluation suite constructed from task-specific skill trajectories, SkillCore preserves skill invocation at rates comparable to those of the corresponding base models. Third, the resulting models maintain or improve performance on the auxiliary SWE-bench Verified evaluation~\citep{jimenez2024swebench,openai2024swebenchverified}. Ablations further show that retaining generalizable-skill scaffolding leads to more runtime skill invocations when skill content is unavailable, whereas omitting skill-native supervision suppresses appropriate task-specific calls.

% \paragraph{Contributions.}
% \begin{itemize}
%   \item We formulate skill internalization as transferring reusable procedural capabilities from runtime skill scaffolds into the model while preserving the ability to access task-specific external information when needed.
%   \item We introduce SkillCore, a framework that integrates a faithful trajectory construction pipeline with a tailored supervised fine-tuning recipe for skill internalization.
%   \item On Qwen3-8B and Qwen2.5-72B, SkillCore improves response quality in the skill-free setting by $54.4\%$ and $24.8\%$, respectively, while preserving task-specific skill invocation capabilities and maintaining or improving general coding performance on SWE-bench Verified.
% \end{itemize}

% !TEX root = ../AnonymousSubmission2027.tex
\section{Related Work}

\begin{figure*}[t]
    \centering
    \includegraphics[width=1\linewidth]{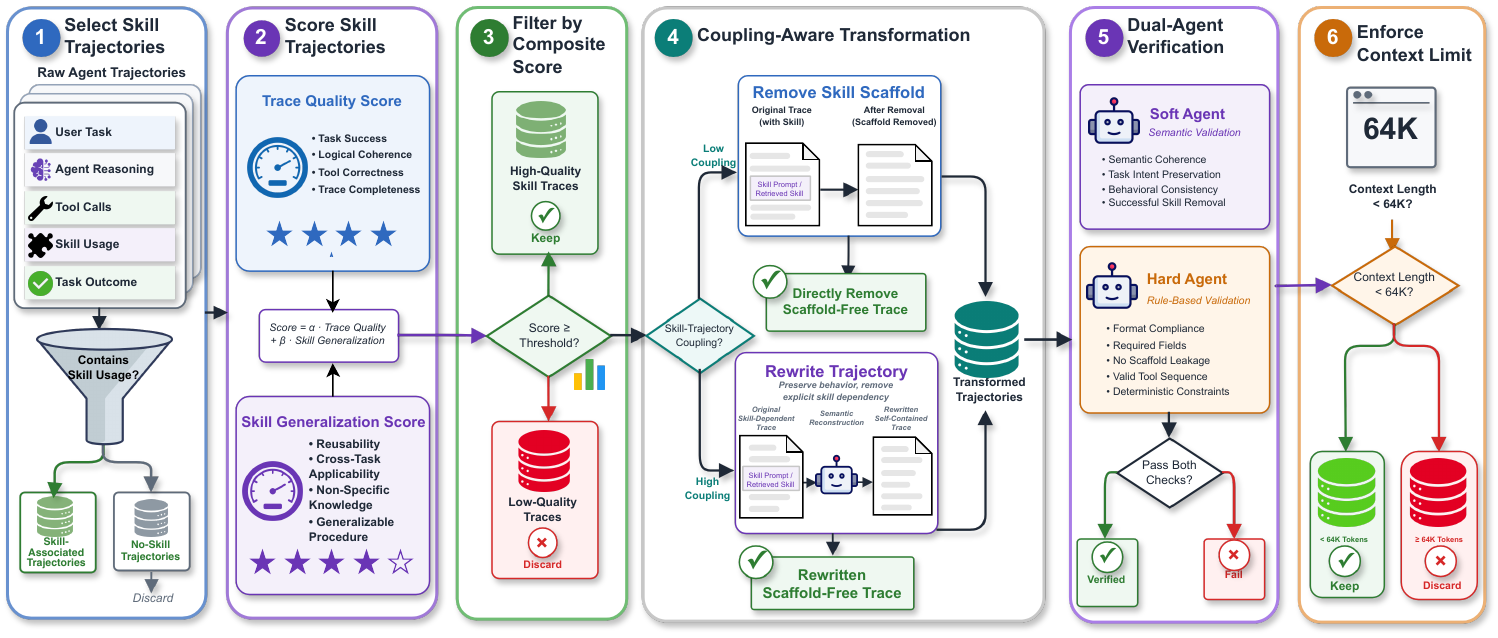}
    \caption{SkillCore's trajectory curation and transformation pipeline. We select skill-using trajectories, score them for quality and generalizability, and keep the high-quality ones with generalizable skills. Each is then transformed to remove the skill scaffold, validated for structural and semantic faithfulness, and filtered by context length to form the scaffold-removed training corpus.}
    \label{fig:pipeline}
\end{figure*}

\subsection{Agent Harnesses and Runtime Skills}
An agent's capabilities depend not only on its underlying model but also on its harness, \emph{i.e.}, the surrounding system of prompts, memory, tools, retrieval mechanisms, and control flow~\citep{sumers2024coala,schluntz2024agents}. For example, ReAct interleaves reasoning with actions in an agent control loop~\citep{yao2023react}, whereas Toolformer and ToolLLM train models to decide when and how to invoke external tools and APIs~\citep{schick2023toolformer,qin2024toolllm}. % Domain-specific interface design can also materially affect agent performance, as demonstrated by the agent--computer interface in SWE-agent~\citep{yang2024sweagent}.
Together, these mechanisms augment the model's parametric capabilities with external information and actions.

Agent skills augment agents at runtime by packaging instructions, scripts, and resources into discoverable directories loaded when relevant~\citep{anthropic2025skills}. Beyond expert-created skills, prior work learns reusable procedures from experience. Voyager learns executable programs through open-ended exploration and retrieves them by natural-language descriptions~\citep{wang2023voyager}, while Agent Workflow Memory abstracts successful trajectories into reusable workflows~\citep{wang2025awm}. Building on these foundations, newer work structures skill acquisition more explicitly: SkillWeaver practices web tasks to synthesize reusable APIs~\citep{zheng2025skillweaver}, SKILL-DISCO compiles shared control-flow subgraphs into executable skills~\citep{guo2026skilldisco}, and SkillRL co-evolves a hierarchical skill bank and agent policy through reinforcement learning~\citep{xia2026skillrl}. Although their representations and acquisition strategies differ, these methods all rely on external skill or workflow libraries retrieved at inference time. SkillCore instead internalizes generalizable procedures to reduce this dependence, while retaining runtime access to task-specific skills.

\subsection{Context Distillation and Skill Internalization}
Context distillation transfers behavior induced by privileged runtime context into model parameters. Prior work distills fixed prompts, instructions, demonstrations, and explanations so models can reproduce the corresponding behavior with less context~\citep{askell2021general,choi2022promptinjection,snell2022contextdistill}. A related line of work distills the results of costly inference-time reasoning into model parameters, allowing direct responses without repeating the reasoning process~\citep{yu2024system2}. Despite demonstrating effective context-to-parameter transfer, these approaches primarily operate on non-interactive input--output examples rather than multi-turn agent trajectories with tool use and environment feedback.

Recent work extends this paradigm to agent trajectories and skills. On-Policy Context Distillation trains a student without privileged context to match a context-conditioned teacher on student-generated trajectories~\citep{ye2026opcd}, while Skill-SD conditions only the teacher on skills summarized from completed on-policy trajectories~\citep{wang2026skillsd}. SKILL0 and SIRI instead integrate skill internalization into agentic reinforcement learning: the former progressively withdraws skill context~\citep{lu2026skill0}, whereas the latter mines and validates skills from current-policy rollouts before distilling beneficial skill-guided actions into the plain policy~\citep{he2026siri}. Together, these approaches depend on newly sampled on-policy behavior, typically obtained from controlled, executable benchmark tasks. LatentSkill offers an offline alternative by converting textual skills into modular LoRA adapters that must still be selected and loaded at inference time~\citep{hu2022lora,yu2026latentskill}. SkillCore instead constructs supervision offline from heterogeneous real-world user trajectories whose original environments cannot be readily reconstructed: it removes generalizable-skill scaffolding while preserving subsequent agent actions, then applies supervised fine-tuning. It therefore requires neither fresh policy rollouts nor an executable environment during corpus construction or training, and internalizes reusable procedures directly into the model's parameters while retaining runtime access to task-specific skills.

% !TEX root = ../AnonymousSubmission2027.tex
\section{Methodology}
\label{sec:method}

\subsection{Overview}
Let $\tau=(\mu_1,\ldots,\mu_L)$ denote an agent trajectory. Each message $\mu_i=(r_i,c_i)$ has one of four roles $r_i$: system, user, assistant, or tool; its content $c_i$ may be natural language, a structured tool call, or a tool result. % SkillCore operates at the level of individual skill calls, so calls within the same trajectory may receive different treatments.
The \emph{skill scaffold} comprises the invocation and returned skill content, assistant utterances whose function is to announce or manage the skill use, and later accesses to the skill's internal files. For generalizable skill calls, we remove this scaffold while preserving the task context, non-target skill interactions, and substantive downstream agent behavior, yielding a scaffold-removed trajectory $\widetilde{\tau}$. The validated outputs form the scaffold-removed corpus $\mathcal{D}_{\mathrm{removed}}$.

Task-specific skills encode procedures or resources that transfer less readily across tasks, making their internalization less broadly useful. Some also depend on external information, proprietary interfaces, or environment-specific functionality that cannot be supplied by model parameters alone. For these skills, we retain \emph{skill-native trajectories}, including their skill calls and returned content, to preserve supervision for runtime invocation. These trajectories constitute the skill-native corpus $\mathcal{D}_{\mathrm{native}}$. SkillCore trains on a mixture of $\mathcal{D}_{\mathrm{removed}}$ and $\mathcal{D}_{\mathrm{native}}$ to balance direct execution of generalizable procedures with runtime use of task-specific skills. Figure~\ref{fig:overview} summarizes the framework; the following subsections describe data construction, training, and evaluation.

\subsection{Data Curation and Scaffold Removal}

\paragraph{Skill-Use Filtering.}
We collect $3{,}352{,}057$ real-world agent trajectories and filter for those containing valid structured \texttt{use\_skill} calls. We consider a call effective only when it specifies a non-empty skill name, is paired with a successful skill-loading result, and is followed by at least one agent action. Only $108{,}514$ trajectories ($3.2\%$) pass this filter, indicating that successfully loaded and subsequently used skills are sparse in the collected trajectories. Because annotating and transforming the entire filtered pool is costly, we select $44{,}960$ trajectories for subsequent annotation and scaffold transformation while preserving coverage over distinct skills.
% A conversation may yield multiple trajectory snapshots with nested or branching interaction histories. We retain the eligible snapshot with the most complete interaction history and an additional branch only if it contains a successfully loaded skill not covered by that conversation's primary snapshot. The selected data comprise $44{,}960$ retained snapshots from $40{,}023$ distinct conversations.

\paragraph{Quality and Generalizability Annotation.}
For each selected trajectory $\tau$, let $\mathcal{C}_{\mathrm{eff}}(\tau)$ denote its set of effective skill calls. We use Claude Opus 4.8~\citep{anthropic2026opus48} to assign a trajectory-level quality score $q(\tau)\in\{1,\ldots,5\}$ and, for each $\kappa\in\mathcal{C}_{\mathrm{eff}}(\tau)$, a generalizability score $\operatorname{gen}(\kappa)\in\{1,\ldots,5\}$ that assesses the reusability of the loaded skill procedure beyond the current task. Scores near $1$ indicate task-specificity, whereas scores near $5$ indicate broad reusability. The judge also returns two Boolean labels: $\operatorname{applied}(\kappa)$ indicates whether subsequent agent actions apply the loaded skill's procedure, and $\operatorname{skillobj}(\kappa)$ indicates whether the user's task targets the skill itself.

Let $\operatorname{validend}(\tau)$ indicate that the task is completed or ends because of collection truncation, $\operatorname{noloop}(\tau)$ that no repeated error loop occurs, and $\operatorname{nodissat}(\tau)$ that the user expresses no dissatisfaction. We define the Boolean trajectory-quality gate and the set of generalizable calls targeted for scaffold removal as
\begin{equation}
\begin{aligned}
\operatorname{pass}(\tau)
&=\mathbb{I}\!\left[q(\tau)\geq4\land\operatorname{validend}(\tau)\right.\\[-2pt]
&\hspace{13mm}\left.\land\operatorname{noloop}(\tau)\land\operatorname{nodissat}(\tau)\right],\\
\mathcal{G}(\tau)
&=\{\kappa\in\mathcal{C}_{\mathrm{eff}}(\tau):\operatorname{gen}(\kappa)\geq4\\[-2pt]
&\hspace{13mm}\land\operatorname{applied}(\kappa)\land\neg\operatorname{skillobj}(\kappa)\}.
\end{aligned}
\end{equation}
where $\mathbb{I}(\cdot)$ denotes the indicator function. Thus, $\operatorname{pass}(\tau)=1$ means that $\tau$ passes the trajectory-quality gate, while $\mathcal{G}(\tau)$ contains the calls selected for scaffold removal. We transform $\tau$ only when $\operatorname{pass}(\tau)=1$ and $\mathcal{G}(\tau)\neq\varnothing$. High-quality trajectories with no call selected for removal remain intact as skill-native supervision.

\paragraph{Coupling-Aware Scaffold Transformation.}
For each trajectory selected for scaffold removal, we construct a compact, indexed representation that fits within Claude Opus 4.8's context window. Messages and skill calls receive stable identifiers, while long system fields and tool results retain only informative prefixes and suffixes. We choose between exact deletion and local rewriting based on whether the target scaffold can be isolated from assistant content. When it can, Claude Opus 4.8 identifies the target calls $\kappa\in\mathcal{G}(\tau)$ and their corresponding text spans, which are then removed deterministically. Otherwise, Claude rewrites only the affected local window and leaves the remainder of the trajectory unchanged. Both routes remove the scaffold associated with calls in $\mathcal{G}(\tau)$, including later accesses to skill-internal files, while preserving non-target skill interactions and downstream agent behavior. We denote the resulting trajectory by $\widetilde{\tau}$.

\paragraph{Transformation Validation.}
We retain a transformed trajectory $\widetilde{\tau}$ only after it passes two complementary validation stages. First, a deterministic structural gate verifies that the invocation and linked result for every target call $\kappa\in\mathcal{G}(\tau)$ have been removed, while all non-target skill calls are retained. It also verifies that all remaining tool calls and results are correctly paired, the system message remains unchanged, and the user request is non-empty. These checks ensure that scaffold removal does not corrupt the agent interaction protocol. However, they cannot determine whether the target scaffold has been fully removed or task-relevant content has been inadvertently deleted.

We then use Claude Opus 4.8 to conduct a semantic review of each transformation that passes the structural gate. It checks six properties: complete removal of the target scaffold, preservation of the original task intent, preservation of task-relevant agent actions, coherence of the edited message sequence, absence of later access to the removed skill's internal files, and continuity of the overall discourse and task structure. A trajectory is retained only if all six checks pass, yielding $10{,}037$ validated scaffold-removed candidates. Figure~\ref{fig:pipeline} summarizes the full data-construction pipeline.

\begin{table*}[t]
\centering
\small
\setlength{\tabcolsep}{3.8pt}
\begin{tabular*}{\textwidth}{@{\extracolsep{\fill}}llccccccc@{}}
\toprule
& & \multicolumn{3}{c}{Task-specific routing} & \multicolumn{3}{c}{Skill-content-free continuation} & Coding \\
\cmidrule(lr){3-5}\cmidrule(lr){6-8}\cmidrule(l){9-9}
Model & Training & Call & Over-int.$\downarrow$ & Corr@Call$\uparrow$ & Win$\uparrow$ & $\Delta$Quality$\uparrow$ & Residual$\downarrow$ & SWE-b$\uparrow$ \\
\midrule
\multirow{4}{*}{Qwen3-8B}
 & Base                      & 50.0 & 50.0 & 89.5 & --   & --               & 16.7 & 4.2 \\
 & Scaffold-retaining        & --   & --   & --   & 58.9 & +0.42 (+16.4\%)  & 44.5 & 5.2 \\
 & Internalization only      & 13.2 & 86.8 & 93.3 & 82.8 & +1.23 (+54.4\%)  & 2.6 & 4.4 \\
 & SkillCore                 & 52.6 & 46.5 & 91.7 & 75.9 & +1.00 (+42.8\%)  & 19.5 & 6.2 \\
\midrule
\multirow{4}{*}{Qwen2.5-72B}
 & Base                      & 65.4 & 28.5 & 98.7 & --   & --                & 33.6 & 43.4 \\
 & Scaffold-retaining        & --   & --   & --   & 35.1 & $-0.37$ ($-11.7\%$) & 59.5 & 45.6 \\
 & Internalization only      & 7.0  & 89.9 & 100.0 & 62.4 & +0.45 (+16.0\%)   & 0.9 & 44.8 \\
 & SkillCore                 & 65.8 & 30.3 & 95.3 & 54.9 & +0.22 (+7.3\%)    & 30.2 & 45.0 \\
\bottomrule
\end{tabular*}
\caption{Main results across the base model and three fine-tuning settings. \emph{Scaffold-retaining} preserves the generalizable-skill scaffolds, whereas \emph{Internalization only} trains exclusively on $\mathcal{D}_{\mathrm{removed}}$. $\Delta$Quality reports absolute and relative changes in the mean 1--5 judge score against the paired base continuation.}
\label{tab:joint-main}
\end{table*}

\subsection{Training}
Before training, we serialize each trajectory using the Hermes agent template implemented in SWIFT~\citep{zhao2025swift}. We estimate serialized sequence lengths with the Qwen2.5-72B tokenizer and retain trajectories below a $65{,}536$-token threshold. % Because this estimate excludes chat-template control tokens, the trainer retains left truncation as a fallback.
We then train on a mixture of $\mathcal{D}_{\mathrm{removed}}$ and $\mathcal{D}_{\mathrm{native}}$, where skill-native trajectories preserve task-specific skill calls and their returned skill content. By default, we set the native-to-removed data ratio to $25\%$. This mixture provides supervision for both direct execution of internalized procedures and runtime invocation of task-specific skills.

We apply loss only to assistant language and structured tool calls. For serialized tokens $x_{1:N}$, let $w_t\in\{0,1\}$ be the loss mask, with $w_t=1$ for assistant tokens and $w_t=0$ otherwise. System, user, and tool-result tokens therefore provide context without supervision:
\begin{equation}
\mathcal{L}(\theta)=-\frac{1}{\sum_{t=1}^{N}w_t}
\sum_{t=1}^{N}w_t\log p_{\theta}(x_t\mid x_{<t}).
\end{equation}

\subsection{Evaluation}
\label{sec:eval}
\paragraph{Skill-Content-Free Continuation.}
This suite tests whether the model can complete tasks supported by a generalizable skill when its content is no longer provided. We construct $348$ continuation tasks spanning $274$ skills, prioritizing broad skill coverage. The suite includes target skills both represented in and absent from the final training corpus, enabling us to assess transfer beyond skills observed during training. We manually review each task to verify the source trajectory, skill categorization, and the validity of the evaluation prefix. Each task provides the base and fine-tuned checkpoints with the same trajectory prefix ending immediately before the target skill call; neither the call nor its returned skill content is provided. Both checkpoints generate up to three assistant rounds, stopping once a round contains no tool call. When a generated round invokes a tool, we replay the next aligned source-recorded tool result rather than execute the tool, so the protocol measures controlled continuation rather than closed-loop success. We evaluate each continuation along two axes. The first is the residual \texttt{use\_skill} invocation rate, which measures continued runtime skill invocation when the target skill content is withheld. The second is overall trajectory quality and task completion, which we assess with an LLM-as-a-judge protocol~\citep{zheng2023judging} using Claude Opus 4.8; it produces a pairwise preference between the base and fine-tuned continuations, together with a $1$--$5$ score for each (higher is better).

\paragraph{Task-Specific Skill Routing.}
This suite tests whether fine-tuning preserves the model's propensity to invoke task-specific skills when they remain necessary at runtime. We construct $228$ evaluation tasks spanning $188$ task-specific skills that depend on external interfaces, private information, or task-specific resources. Each task uses a prefix from a high-quality trajectory whose recorded continuation invokes, successfully loads, and applies the target skill. The prefix ends immediately before the recorded skill call. We manually review each task to verify the source trajectory, skill categorization, and prefix validity. Each model then generates a single next action. We report three metrics that characterize this routing decision. The \emph{call rate} is the fraction of tasks on which the model invokes \texttt{use\_skill}; its change relative to the base checkpoint indicates whether fine-tuning shifts the model's propensity to route through skills. The \emph{over-internalization rate} is the fraction of tasks on which the model instead answers directly or invokes another tool. An increase indicates a shift toward handling the task without invoking the recorded target skill. Among generated \texttt{use\_skill} calls, the \emph{correct-given-called rate} measures how often the model selects the recorded target skill, separating whether to invoke from which skill to select. The call and over-internalization rates use all $228$ tasks as the denominator; empty generations count as non-calls. This evaluates routing before skill loading, not subsequent task success.

\paragraph{General Coding Performance.}
We use SWE-bench Verified~\citep{jimenez2024swebench,openai2024swebenchverified} as an auxiliary benchmark to assess whether fine-tuning preserves general coding performance.

\FloatBarrier

% !TEX root = ../AnonymousSubmission2027.tex
\section{Experiments}
\label{sec:experiments}

\subsection{Overview}
We evaluate SkillCore with Qwen3-8B and an agent-post-trained Qwen2.5-72B checkpoint on two complementary suites: skill-content-free continuation, which measures task performance when target skill content is withheld, and task-specific skill routing. We ask whether scaffold-removed supervision improves task completion while reducing residual skill invocation (RQ1); whether the observed performance gains arise from scaffold removal rather than training on the same trajectories with scaffolding retained (RQ2); how internalization-only training affects task-specific skill routing (RQ3); and how skill-native supervision trades off routing preservation against internalization (RQ4). We additionally examine the effects of corpus construction choices and check general coding performance.

\subsection{Experimental Settings}
All models undergo full-parameter fine-tuning using the Megatron backend~\citep{narayanan2021megatron} and Hermes agent template implemented in ms-swift~\citep{zhao2025swift}. We use a maximum sequence length of $65{,}536$ tokens, with left truncation retained only as a fallback after template serialization. System, user, and tool-result messages provide context without contributing to the loss, whereas assistant language and tool calls are supervised. We use Adam with an initial learning rate of $10^{-5}$, cosine decay to $10^{-6}$, a $5\%$ warmup ratio, weight decay of $0.1$, gradient clipping at $1.0$, and seed 42. The global batch size is 128 for 8B and 64 for 72B, with a microbatch size of 1. We train the 8B and 72B models for 6 and 3 epochs, respectively. Training uses two nodes, each with eight 141-GB NVIDIA H200 GPUs.

\subsection{RQ1: Can Scaffold-Removed Supervision Internalize Generalizable Skills?}
We evaluate \emph{internalization-only} models trained exclusively on $\mathcal{D}_{\mathrm{removed}}$, without skill-native data from $\mathcal{D}_{\mathrm{native}}$. Table~\ref{tab:joint-main} reports their performance on the skill-content-free continuation suite. For Qwen3-8B, the residual skill-invocation rate falls from $16.7\%$ to $2.6\%$; its continuations win $82.8\%$ of pairwise comparisons against those from the base checkpoint, while the mean judge score increases by $1.23$ points ($54.4\%$). For Qwen2.5-72B, residual invocation falls from $33.6\%$ to $0.9\%$; its continuations win $62.4\%$ of comparisons, with a mean judge-score increase of $0.45$ points ($16.0\%$). Across both model scales, scaffold-removed supervision improves task completion quality while substantially reducing residual skill invocation.
% Across the 8B training-duration sweep, quality peaks at epoch~6 while residual invocation stays near $3\%$ throughout (see Appendix).

\subsection{RQ2: Is Scaffold Removal Necessary?}
The \emph{Scaffold-retaining} control uses the same source trajectories and training epochs as internalization-only, but retains the generalizable-skill scaffolds. Compared with scaffold removal, retaining them raises residual invocation from $2.6\%$ to $44.5\%$ for 8B and from $0.9\%$ to $59.5\%$ for 72B, while reducing pairwise win rates from $82.8\%$ to $58.9\%$ and from $62.4\%$ to $35.1\%$, respectively. It also yields smaller quality changes: $+0.42$ points ($+16.4\%$) rather than $+1.23$ ($+54.4\%$) for 8B, and $-0.37$ ($-11.7\%$) rather than $+0.45$ ($+16.0\%$) for 72B. Overall, this matched comparison indicates that the gains arise from scaffold removal rather than fine-tuning alone, enabling task completion with far less reliance on external skills.

\subsection{RQ3: What Routing Cost Does Internalization-Only Training Impose?}
The gains on generalizable skills come at a substantial cost to task-specific skill routing (Table~\ref{tab:joint-main}). On the task-specific routing suite, the call rate drops from $50.0\%$ to $13.2\%$ for 8B and from $65.4\%$ to $7.0\%$ for 72B. Correspondingly, the over-internalization rate rises from $50.0\%$ to $86.8\%$ for 8B and from $28.5\%$ to $89.9\%$ for 72B, as the models respond directly or invoke another tool instead of calling the target skill. Among the relatively few generated skill calls, however, the target skill is usually selected correctly (Corr@Call: $93.3\%$ and $100\%$). The main degradation therefore lies in deciding whether to invoke a skill, not in selecting which skill to invoke once a call is made. Overall, internalization-only training substantially weakens task-specific skill routing despite preserving conditional selection accuracy.

\subsection{RQ4: Can Skill-Native Supervision Recover Task-Specific Routing?}
We vary the native-to-removed ratio from $0\%$ to $50\%$, where mixX denotes an X\% ratio (Table~\ref{tab:mixture-full}). Increasing this ratio consistently raises task-specific call rates and reduces over-internalization, but also increases residual invocation and generally weakens skill-content-free continuation performance. From no mix to mix50, call rates rise from $13.2\%$ to $66.7\%$ for 8B and from $7.0\%$ to $73.2\%$ for 72B, while residual invocation rises from $2.6\%$ to $27.0\%$ and from $0.9\%$ to $32.2\%$, respectively. Thus, skill-native supervision recovers routing at the cost of greater dependence on external skills. We use mix25 as the default SkillCore configuration because it most closely matches the base models' call rates ($52.6\%$ vs.\ $50.0\%$ for 8B; $65.8\%$ vs.\ $65.4\%$ for 72B), while retaining win rates above $50\%$ and positive judge-score gains.

\begin{table}[t]
\centering
\small
\setlength{\tabcolsep}{3.2pt}
\renewcommand{\arraystretch}{1.0}
\begin{tabular}{@{}lrrrr@{}}
\toprule
Metric & No mix & mix10 & mix25 & mix50 \\
\midrule
\multicolumn{5}{@{}l}{\emph{Qwen3-8B}} \\
Call & 13.2\% & 44.3\% & 52.6\% & 66.7\% \\
Over-int.$\downarrow$ & 86.8\% & 55.3\% & 46.5\% & 32.9\% \\
Corr@Call$\uparrow$ & 93.3\% & 92.1\% & 91.7\% & 88.2\% \\
\addlinespace[1pt]
Win$\uparrow$ & 82.8\% & 75.9\% & 75.9\% & 74.7\% \\
$\Delta$Quality$\uparrow$ & +1.23 & +0.95 & +1.00 & +0.91 \\
Residual$\downarrow$ & 2.6\% & 12.4\% & 19.5\% & 27.0\% \\
\addlinespace[1pt]
SWE-b$\uparrow$ & 4.4\% & 4.8\% & 6.2\% & 8.6\% \\
\midrule
\multicolumn{5}{@{}l}{\emph{Qwen2.5-72B}} \\
Call & 7.0\% & 60.1\% & 65.8\% & 73.2\% \\
Over-int.$\downarrow$ & 89.9\% & 36.4\% & 30.3\% & 24.1\% \\
Corr@Call$\uparrow$ & 100.0\% & 94.9\% & 95.3\% & 93.4\% \\
\addlinespace[1pt]
Win$\uparrow$ & 62.4\% & 60.1\% & 54.9\% & 54.9\% \\
$\Delta$Quality$\uparrow$ & +0.45 & +0.45 & +0.22 & +0.23 \\
Residual$\downarrow$ & 0.9\% & 21.6\% & 30.2\% & 32.2\% \\
\addlinespace[1pt]
SWE-b$\uparrow$ & 44.8\% & 44.6\% & 45.0\% & 46.8\% \\
\bottomrule
\end{tabular}
\caption{Routing--internalization trade-off across skill-native mixture ratios. No mix denotes internalization-only training, and mix25 is the default SkillCore configuration. $\Delta$Quality denotes absolute judge-score changes.}
\label{tab:mixture-full}
\end{table}

\begin{figure*}[t]
\centering
\begin{tikzpicture}
\definecolor{ftblue}{HTML}{1478B8}
\definecolor{basegray}{HTML}{737B85}
\definecolor{negorange}{HTML}{D97706}
\definecolor{zerogray}{HTML}{A8AFB7}

\begin{axis}[
  name=frequency,
  width=0.49\textwidth,
  height=4.45cm,
  title={(a) Quality across training-frequency groups},
  title style={font=\small,yshift=-1pt},
  xmin=-0.35, xmax=5.35,
  ymin=1.9, ymax=4.05,
  xtick={0,1,2,3,4,5},
  xticklabels={
    {\shortstack{0\\[-1pt]\scriptsize $(n{=}152)$}},
    {\shortstack{1--2\\[-1pt]\scriptsize $(n{=}131)$}},
    {\shortstack{3--5\\[-1pt]\scriptsize $(n{=}48)$}},
    {\shortstack{6--10\\[-1pt]\scriptsize $(n{=}13)$}},
    {\shortstack{11--20\\[-1pt]\scriptsize $(n{=}14)$}},
    {\shortstack{21+\\[-1pt]\scriptsize $(n{=}23)$}}
  },
  xlabel={Target-skill frequency in the final training corpus},
  ylabel={Mean judge score},
  tick label style={font=\footnotesize},
  label style={font=\footnotesize},
  legend style={
    font=\footnotesize,
    draw=none,
    fill=none,
    at={(0.5,0.98)},
    anchor=north,
    legend columns=2,
    /tikz/every even column/.append style={column sep=8pt}
  },
  axis line style={black!75},
  tick style={black!75},
  ymajorgrids,
  grid style={black!12},
  clip=false,
]
  \addplot[forget plot,draw=ftblue,fill=ftblue!10] coordinates {(-0.35,1.9) (-0.35,4.05) (0.35,4.05) (0.35,1.9)} \closedcycle;
  \addplot+[ftblue,very thick,mark=*,mark size=2.7pt,mark options={fill=ftblue,draw=black!65}] coordinates {
    (0,3.47) (1,3.44) (2,3.71) (3,3.32) (4,3.50) (5,3.70)
  };
  \addlegendentry{Fine-tuned}
  \addplot+[basegray,thick,densely dashed,mark=square*,mark size=2.5pt,mark options={fill=white,draw=basegray}] coordinates {
    (0,2.28) (1,2.27) (2,2.25) (3,2.23) (4,2.36) (5,2.13)
  };
  \addlegendentry{Base}
\end{axis}

\begin{axis}[
  at={(frequency.east)},
  anchor=west,
  xshift=1.85cm,
  width=0.42\textwidth,
  height=4.45cm,
  title={(b) Distribution of paired score changes},
  title style={font=\small,yshift=-1pt},
  xmin=-3.6, xmax=4.6,
  ymin=0, ymax=215,
  xtick={-3,-2,-1,0,1,2,3,4},
  ytick={0,50,100,150,200},
  xlabel={Score change after fine-tuning},
  ylabel={Tasks},
  tick label style={font=\footnotesize},
  label style={font=\footnotesize},
  axis line style={black!75},
  tick style={black!75},
  ymajorgrids,
  grid style={black!12},
  clip=false,
]
  \addplot[ybar,bar width=13pt,draw=black!75,fill=negorange] coordinates {(-3,6) (-2,20) (-1,29)};
  \addplot[ybar,bar width=13pt,draw=black!75,fill=zerogray] coordinates {(0,11)};
  \addplot[ybar,bar width=13pt,draw=black!75,fill=ftblue] coordinates {(1,73) (2,189) (3,17) (4,3)};
  \node[font=\scriptsize,anchor=south] at (axis cs:-3,6) {6};
  \node[font=\scriptsize,anchor=south] at (axis cs:-2,20) {20};
  \node[font=\scriptsize,anchor=south] at (axis cs:-1,29) {29};
  \node[font=\scriptsize,anchor=south] at (axis cs:0,11) {11};
  \node[font=\scriptsize,anchor=south] at (axis cs:1,73) {73};
  \node[font=\scriptsize,anchor=south] at (axis cs:2,189) {189};
  \node[font=\scriptsize,anchor=south] at (axis cs:3,17) {17};
  \node[font=\scriptsize,anchor=south] at (axis cs:4,3) {3};
\end{axis}
\end{tikzpicture}
\caption{Quality analysis for the selected Qwen3-8B internalization-only checkpoint. (a) Mean judge scores of the fine-tuned and base models across target-skill frequency groups; $n$ denotes evaluated task--skill pairs. (b) Number of tasks by paired judge-score change after fine-tuning.}
\label{fig:transfer-quality}
\end{figure*}
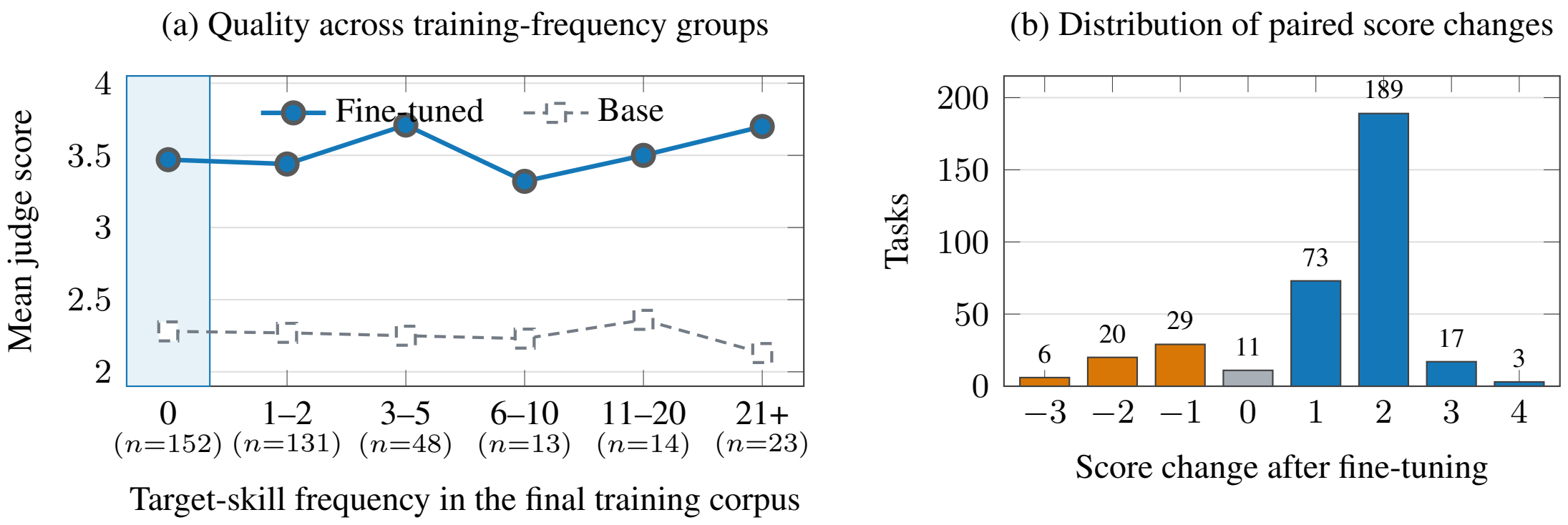

\subsection{Data Scalability}
Starting from the core scaffold-removed corpus, we construct two doubled variants. The first adds trajectories whose serialized lengths fit within the 64K context limit, preserving their full content. The second adds trajectories exceeding 64K tokens and left-truncates them to the limit during training. Table~\ref{tab:data-curation} compares these variants. Data scaling has little effect on Qwen3-8B: the core and both expanded variants achieve similar win rates ($81.3$--$82.8\%$) and quality gains ($53.1$--$54.4\%$). Qwen2.5-72B benefits more clearly. The $\leq$64K variant raises the win rate from $62.4\%$ to $71.3\%$ and the quality gain from $16.0\%$ to $24.8\%$. The left-truncated variant yields smaller improvements ($64.7\%$ win and $18.9\%$ quality gain), suggesting that truncation may remove task-relevant context. Both expanded variants increase residual skill invocation, so the additional data improves continuation quality without further reducing runtime skill dependence.

% Across Tables~\ref{tab:joint-main} and~\ref{tab:mixture-full}, fine-tuning preserves or improves SWE-bench Verified performance. For Qwen3-8B, all fine-tuned settings exceed the $4.2\%$ base score, reaching $6.2\%$ for SkillCore and $8.6\%$ at mix50. For Qwen2.5-72B, scores remain above the $43.4\%$ base and range from $44.6\%$ to $46.8\%$ across mixture ratios. Thus, neither scaffold removal nor the skill-native mixture degrades general coding capability.

% !TEX root = ../AnonymousSubmission2027.tex
\section{Analysis}
\label{sec:analysis}

\paragraph{Transfer to Skills Absent from the Final Training Corpus.}
Because some evaluation tasks involve multiple target skills, the $348$ tasks correspond to $381$ task--skill pairs. Among them, $152$ pairs cover $120$ target skills absent from the final length-filtered training corpus. For the selected Qwen3-8B internalization-only checkpoint, this zero-frequency group achieves $82.2\%$ pairwise wins, a mean judge score of $3.47$, and $2.0\%$ residual invocation. Figure~\ref{fig:transfer-quality} shows consistent quality gains across all frequency groups, including skills absent from the final training corpus. The gains do not increase monotonically with training frequency.

\paragraph{Paired Quality Gains.}
The paired distribution in Figure~\ref{fig:transfer-quality} shows broad improvement rather than gains driven by a few outliers. Among the $348$ skill-content-free tasks, $282$ improve, $11$ remain unchanged, and $55$ decline. The most common change is a two-point increase, observed on $189$ tasks, and the mean change is $+1.23$. These results indicate a systematic quality improvement.

\begin{table}[t]
\centering
\small
\setlength{\tabcolsep}{3pt}
\resizebox{\columnwidth}{!}{%
\begin{tabular}{lccc}
\toprule
Corpus & Win$\uparrow$ & $\Delta$Quality$\uparrow$ & Residual$\downarrow$ \\
\midrule
\multicolumn{4}{l}{\emph{Qwen3-8B}} \\
Core & 82.8\% & +1.23 (+54.4\%) & 2.6\% \\
Expanded ($\leq$64K) & 82.2\% & +1.21 (+53.1\%) & 6.0\% \\
Expanded ($>$64K, trunc.) & 81.3\% & +1.21 (+54.0\%) & 5.5\% \\
\midrule
\multicolumn{4}{l}{\emph{Qwen2.5-72B}} \\
Core & 62.4\% & +0.45 (+16.0\%) & 0.9\% \\
Expanded ($\leq$64K) & 71.3\% & +0.70 (+24.8\%) & 7.5\% \\
Expanded ($>$64K, trunc.) & 64.7\% & +0.54 (+18.9\%) & 12.1\% \\
\bottomrule
\end{tabular}
}
\caption{Data scalability results for the core corpus and two doubled variants.}
% $\Delta$Quality reports absolute and relative changes in mean judge score over the paired base model.
\label{tab:data-curation}
\end{table}

\paragraph{Judge Consistency.}
We check whether the pairwise preferences used for the win rate agree with the separately assigned quality scores. In $97.9$--$98.2\%$ of examples, the preferred continuation also receives the higher score, and $67.5\%$ of comparisons differ by at least two points. A 50-case position-swap audit further shows that changing response order affects judgments mainly when the two scores are close. Thus, the reported win rates are well supported by the scalar quality scores and are not primarily driven by response order.

% \input{tables/tab_judge_checks}

% !TEX root = ../AnonymousSubmission2027.tex
\section{Conclusion}

Motivated by the low utilization of runtime skills, we argue that generalizable skills should be internalized while task-specific skills providing external information remain callable. We present SkillCore, which removes generalizable-skill scaffolds while preserving task descriptions and subsequent actions, and mixes in a small proportion of skill-native trajectories to retain task-specific skill invocation. SkillCore improves the skill-free response quality of Qwen3-8B and Qwen2.5-72B over their base models by $54.4\%$ and $24.8\%$, respectively, while preserving task-specific skill invocation and general coding ability. SkillCore thereby makes capabilities delivered through runtime scaffolds intrinsic and available to all users by default.

% !TEX root = ../NAACL2027.tex
\section*{Limitations}

Our evaluation uses one training seed, one deterministic generation per checkpoint, and one Claude Opus 4.8 judgment per task. We therefore do not estimate variation across training runs, generations, or judges, and report point estimates rather than confidence intervals. The continuation suite replays recorded tool results instead of executing generated actions, while the routing suite evaluates only the decision to invoke a skill. Neither suite measures closed-loop task success.

Because the study uses real-world agent trajectories, sharing the data or derived models requires appropriate access controls and privacy review. Full-parameter fine-tuning at long context lengths also requires substantial computation, which may limit reproducibility and practical adoption. These constraints bound our claims to the evaluated models, data distribution, and protocols.

\FloatBarrier
\bibliography{references}
\FloatBarrier

\end{document}